\RequirePackage{fix-cm}
\documentclass[twocolumn]{svjour3}          % twocolumn
\smartqed  % flush right qed marks, e.g. at end of proof
\usepackage{graphicx}

\usepackage{amsmath,amssymb}
\usepackage{algorithm}
\usepackage{tabularx}
\usepackage{upgreek}
\usepackage{multirow}
\begin{document}

\title{Weighted Least Squares Twin Support Vector Machine with Fuzzy Rough Set Theory for Imbalanced Data Classification}


\author{Maysam Behmanesh  \and
        Peyman Adibi \and
        Hossein Karshenas
}


\institute{Maysam Behmanesh \at
              \email{mbehmanesh@eng.ui.ac.ir}
           \and
           Peyman Adibi (Corresponding author) \at
              \email{adibi@eng.ui.ac.ir}
           \and
           Hossein Karshenas \at
              \email{h.karshenas@eng.ui.ac.ir}
           \and
           	Department of Artificial Intelligence, Faculty of Computer Engineering, University of Isfahan, Iran
}

\date{Received: May 2021 / Accepted: date}

\maketitle

\begin{abstract}
Support vector machines (SVMs) are powerful supervised learning tools developed to solve classification problems. However, SVMs are likely to perform poorly in the classification of imbalanced data. The rough set theory presents a mathematical tool for inference in nondeterministic cases that provides methods for removing irrelevant information from data. In this work, we propose an approach that efficiently used fuzzy rough set theory in weighted least squares twin support vector machine called FRLSTSVM for classification of imbalanced data. The first innovation is introducing a new fuzzy rough set based under-sampling strategy to make the classifier robust in terms of the imbalanced data. For constructing the two proximal hyperplanes in FRLSTSVM, data points from the minority class remain unchanged while a subset of data points in the majority class are selected using a new method. In this model, we embed the weight biases in the LSTSVM formulations to overcome the bias phenomenon in the original twin SVM for the classification of imbalanced data. In order to determine these weights in this formulation, we introduce a new strategy that uses fuzzy rough set theory as the second innovation. Experimental results on the famous imbalanced datasets, compared to the related traditional SVM-based methods, demonstrate the superiority of the proposed FRLSTSVM model in the imbalanced data classification.
\end{abstract}


\section{Introduction} \label{sec.intr}
Nowadays, classification of imbalanced data is very important. Generally, real datasets are formed from large amounts of majority data points and a little percent of minority data points. For example, mammography datasets are containing about 98\% of normal data (majority) and 2\% of abnormal data (minority). In general, these datasets are classified in such a way that the class with higher importance (minority class) has fewer data points than the class with lower importance (majority class). Many various methods are presented for classification of such imbalanced datasets. Several efforts are dedicated to increase the samples of minority class (over-sampling) and decrease the data samples of majority class (under-sampling).\par

Support vector machines (SVM)\cite{Burges1998,Cortes1995,Deng2012} are the popular classifiers recognized as the most useful kernel-based tools in the supervised learning problems, such as classification and regression, and they are successfully used in many real-world problems \cite{Yu-Xin2012,Isa4492780}.\par

Learning complexity is one of the main SVM challenges, because the quadratic programming problem (QPP) of the training phase is difficult to solve and its computational complexity is $O(m^3)$, where $m$ is the total number of the training points \cite{Deng2012}. Twin support vector machine (TSVM), which is a solution to raise the computational speed of SVM, has proposed in recent years \cite{Shao5762620} in binary classification problems. Instead of solving one complex QPP, TSVM tries to solve two smaller QPPs to determine two nonparallel hyperplanes such that each hyperplane is closer to one class and as far as possible from the other one. The size of QPP in TSVM is $m/2$ rather than $m$ in SVM and thus, the optimization procedure of TSVM is faster than the standard SVM.\par

Standard SVM is trained based on the assumption of balanced datasets. When applying SVM on imbalanced datasets, since the data points in majority class are dominant, the learning of minority data points is not performed very well and results in the improper classification for minority class.\par

Fuzzy rough set theory is one of the best methods in dealing with uncertainty in data, and since data in majority class generally contain redundant instances, we used this theory to select samples that provide the best representation of the majority data and also alleviate imbalanced in classification. To this end, we propose a new method for classification of imbalanced data that is called fuzzy rough weighted least squares twin support vector machine (FRLSTSVM). In order to construct two hyperplanes with weighted least squares twin SVM (WLTSVM) in an imbalanced data classification problem, different training points are used, such that minority data points remain unchanged while a subset of the majority data points is selected using a new method based on fuzzy rough set theory. The main contributions of our FRLSTSVM  model are: (1) introduce a new under-sampling strategy based on the fuzzy rough set theory to reduce effectively the number of majority training points and make resistant to the outliers and imbalanced data; (2) overcome the bias phenomenon to majority class in the original TSVM by embedding weight biases in the least squares TSVM formulations by introducing a new strategy based on fuzzy rough set theory to determine and analyze the weight biases in the WLTSVM formulation for improving its performance.\par

The remainder of this paper is organized as follows. Section \ref{2relWor} outlines the related work. Section \ref{3.LeastSquaresTwinSupportVectorMachine} briefly explains SVM and twin SVM classifiers. In section \ref{4RoughSetTheory}, the fuzzy-rough set theory is presented. The proposed method for imbalanced data classification is presented in section \ref{5ProposedMethod}. Finally, obtained results from experiments conducted on several datasets and comparison of the proposed method results with those of several related techniques are presented in section \ref{6ExperimentalResults}. The concluding remarks are given in section \ref{7Conclusions}.

\section{Related work}
\label{2relWor}

In imbalanced data problem, various methods and algorithms are proposed. The weighted SVM and LSSVM approachs are presented in \cite{Yang1555965}. In \cite{Fung2005}, the weight factors are applied to the SVM formulation to handle the imbalanced problems with assigning different weights to the training samples in the majority and minority classes. Also, the weighted TSVM approach for the imbalanced classification is presented in \cite{Jing5451483}, which puts the different weights on the error variables, eliminates the impact of the noise data, and obtains a robust estimation.\par

The data-preprocessing approache is one the main approaches in deal with the imbalanced data that tries to over-sampling in the minority data points \cite{DBLP2011} or under-sampling in the majority data points \cite{Kubat97addressingthe}. Many samples of the preprocessing approaches are presented in \cite{DBLP2011,Kubat97addressingthe,Batista2004}. The main strtegy of these approaches is reducing the impact of the inappropriate data points in learning the classifier, which in many cases may cause to overfitting. Thus, a number of heuristic methods are proposed to introduce a trade-off in this mean.\par

A method is presented in \cite{Kubat97addressingthe} that applies under-sampling on data points in majority class, while preserving the original content of the minority class. In this method, the noisy samples from the majority class and borderline samples, those lying either on or very close to the decision boundaries between classes, are detected and removed using the method described in \cite{10.1007/3-540-48229-6_9}. The redundant majority class samples, majority class samples which are quite distant from the decision border, are also removed. Another approach for detecting and removing the noisy and borderline samples is presented in \cite{4309452}, which removes the majority class samples whose one class is different from the class of at least two of its three nearest neighbors. \par

A graph based strategy under-sampling method is proposed in \cite{SHAO20143158} that is robustness to outliers and overcomes the bias phenomenon in the original TWSVM by embedding the weight biases in the TWSVM formulations. An approach is proposed in \cite{LIU2019702}, which combines easy-tuning version of SVM with a modified feature vector selection techniques for selecting a small number of samples. This approach represents linearly all the dataset and also gives a maximal separability of the imbalanced data in the Reproducing Kernel Hilbert Space (RKHS).\par

Other popular approaches in handling imbalanced data are over-sampling of the minority objects in their neighborhood. One approach is called SMOTE, which over-samples the minority class by creating new synthetic examples. The main idea of this approach is to generate these new data points by interpolating several samples of minority class that are closed to each other \cite{DBLP2011}. The combinations of the SMOTE method and other subsampling methods are presented and hopeful results are reported in \cite{Batista2004}. Radial-Based Oversampling (RBO) method \cite{KOZIARSKI201919} is anoother over-sampling approach, which tries to find regions in which the synthetic objects from the minority class should be generated on the basis of the imbalance distribution estimation with radial basis functions.\par

Another category of the imbalanced data classification methods tries to control the detection rate of the minority class by adding a cost term or decision threshold to the classifier and adjusting it. For example, to learn the optimal decision boundary in the classification problem, the ratio of the majority samples in all the training data are decreased in \cite{Elkan2001,5299216}. Also, in \cite{Chew2000,Joachims99transductiveinference}, the prediction performance of the imbalanced data is improved by tuning a weight for each class. The margin distribution theory is used in \cite{CHENG2016107} to design a balanced classifier by introducing a Cost Sensitive Large margin Distribution Machine (CS-LDM) to improve the detection rate of the minority class by using cost-sensitive margin mean and cost-sensitive penalty.\par

Some approaches for the problem of imbalanced data apply the misclassifications as a loss function to update the distribution of the training data on successive boosting rounds \cite{Fan1999,SUN20073358,5978225}. A robust lost function to deal with the imbalanced problem in the noise environment is designed in \cite{WANG201940}. Applying this loss function into the SVM formulation presents a robust SVM framework which results in a Bayes optimal classifier.\par

Several other combined ways are presented to confront imbalanced data problem such as RUSBoost \cite{5299216} that proposes to combine data sampling methods with fast boosting. In order to ensure more minority instances for successive classifiers, a misclassification cost weights determination method is proposed in \cite{TAO201931}. This method considers the different contribution of minority instances to the form of SVM classifiers at each iteration based on the preceding obtained classifier during boosting, which can allow it to produce diverse classifiers and thus improve its generalization performance.\par

Recently, rough set theory as an important mathematical tool for dealing with the fuzziness and uncertainties in data, has attracted more and more attention in the classification of imbalanced data. An algorithm for online streaming feature selection is proposed in \cite{ZHOU2017187} that uses the information of the nearest neighbors providing by rough sets theory to select the relevant features which can get higher separability between the majority class and the minority class. \par

An approach is proposed in \cite{CHEN20191} that provides feature selection of the imbalanced data by employing the neighborhood rough set theory. In this approach, the significance of the features is defined by carefully studying the upper and lower boundary regions and the uneven distribution of the classes is considered during the definition of the feature significance.\par

To improve the performance of the SMOTE method, an approach is proposed in \cite{VERBIEST2014511} that uses two prototype selection techniques both based on the fuzzy rough set theory. The first fuzzy rough prototype selection algorithm removes noisy instances from the imbalanced dataset, the second cleans the data generated by the SMOTE.\par

An approach for classification of  imbalanced data is introduced in \cite{Ramentol2015} that uses the fuzzy rough set theory and ordered weighted average aggregation by considering different strategies to build a weight vector to take into account data imbalance. \par

In many approaches, fuzzy-rough set theory is applied before the classification methods, which needs considerable computation time. An approach is proposed in \cite{Nguyen2015} that uses different criteria for the minority and majority classes in the fuzzy-rough instance selection and eliminates the time consuming step employed in the previous approaches.

\section{Least Squares Twin Support Vector Machine}
\label{3.LeastSquaresTwinSupportVectorMachine}
Twin support vector machine (TSVM) \cite{5762620} seeks a pair of non-parallel hyperplanes to accelerate the training process of SVM classifier. To more speed up this training process, least squares TSVM (LTSVM) is proposed that also has better generalization capability than TSVM \cite{ARUNKUMAR20097535}.\par

Consider $m_{1}$ positive class data points and $m_{2}$ negative class data points in the $n$-dimensional real space $\mathbb{R}^{n}$, where $m_{1}+m_{2}=m$. Let $\mathbf{X}_{1}\in \mathbb{R}^{m_{1} \times n}$ is a matrix for representing the positive class instances, and $\mathbf{X}_{2}\in \mathbb{R}^{m_{2} \times n}$ is a matrix for representing the negative class instances. In imbalanced classification problems, it is generally assumed that the data points belonging to the positive class are minority and the data points belonging to the negative class are the majority.\par

Two non-parallel hyperplanes in $\mathbb{R}^{n}$ are given below:

\begin{equation}
\begin{aligned}
&f_{1}(\mathbf{x})=\mathbf{w}_{1}^T\mathbf{x}+b_{1}=0,\\
&f_{2}(\mathbf{x})=\mathbf{w}_{2}^T\mathbf{x}+b_{2}=0,
\end{aligned}
\end{equation}

\noindent where $\mathbf{w}\in \mathbb{R}^{n}$ and $b\in \mathbb{R}$ and these hyperplanes are obtained from the solutions of the following primal problem of LTSVM:

\begin{equation}
\label{eq.2}
\begin{aligned}
&\min_{\mathbf{w}_{1},b_{1},\mathbf{\upxi}} \frac{1}{2}\parallel \mathbf{X}_{1}\mathbf{w}_{1}+\mathbf{e}_{1}b_{1}\parallel^{2}+\frac{c_{1}}{2}\mathbf{\upxi}^{T}\mathbf{\upxi},
\\
&s.t. \quad -(\mathbf{X}_{2}\mathbf{w}_{1}+\mathbf{e}_{2}b_{1})+\mathbf{\upxi}\geq \mathbf{e}_{2}, \quad \mathbf{\upxi}\geq 0
\end{aligned}
\end{equation}
and

\begin{equation}
\label{eq.3}
\begin{aligned}
&\min_{\mathbf{w}_{2},b_{2},\mathbf{\upeta}} \frac{1}{2}\parallel \mathbf{X}_{2}\mathbf{w}_{2}+\mathbf{e}_{2}b_{2}\parallel^{2}+\frac{c_{2}}{2}\mathbf{\upeta}^{T}\mathbf{\upeta},
\\
&s.t. \quad (\mathbf{X}_{1}\mathbf{w}_{2}+\mathbf{e}_{1}b_{2})+\mathbf{\upeta}\geq \mathbf{e}_{1}, \quad \mathbf{\upeta}\geq 0
\end{aligned}
\end{equation}

\noindent where $\mathbf{\upxi}$ and $\mathbf{\upeta}$ are two slack variables, $\mathbf{e}_{1}$ and $\mathbf{e}_{2}$ are two vectors with suitable dimensions having all values as $1$s, and  $c_{1}>0$ and $c_{2}>0$ are penalty parameters. Lagrangian forms of equations (\ref{eq.2}) and (\ref{eq.3}) are given as follows:

\begin{equation}
\label{eq.Lag1}
\begin{aligned}
L(\mathbf{w}_{1},b_{1},\mathbf{\upxi},\mathbf{\upalpha})&=\frac{1}{2}\parallel \mathbf{X}_{1}\mathbf{w}_{1}+\mathbf{e}_{1}b_{1}\parallel^{2}+\frac{c_{1}}{2}\mathbf{\upxi}^{T}\mathbf{\upxi}\\
&-\mathbf{\upalpha}^{T}(-(\mathbf{X}_{2}\mathbf{w}_{1}+e_{2}b_{1})+\mathbf{\upxi}-e_{2}),
\end{aligned}
\end{equation}

and
\begin{equation}
\begin{aligned}
L(\mathbf{w}_{2},b_{2},\mathbf{\upeta},\mathbf{\upbeta})&=\frac{1}{2}\parallel \mathbf{X}_{2}\mathbf{w}_{2}+e_{2}b_{2}\parallel^{2}+\frac{c_{2}}{2}\mathbf{\upeta}^{T}\mathbf{\upeta}\\
&-\mathbf{\upbeta}^{T}((\mathbf{X}_{1}\mathbf{w}_{2}+e_{1}b_{2})+\mathbf{\upeta}-\mathbf{e}_{1}),
\end{aligned}
\end{equation}

\noindent where $\mathbf{\upalpha}\in \mathbb{R}^{n}$ and $\mathbf{\upbeta} \in \mathbb{R}^{n}$ are the vectors of Lagrangean multipliers. With KKT conditions, the derivatives of equation (\ref{eq.Lag1}) with respect to $\mathbf{w}_{1}$, $b_{1}$, $\mathbf{\upxi}$ and $\mathbf{\upalpha}$, are calculated and set equal to zero, as follows:

\begin{equation}
\label{eq.difW}
\begin{aligned}
&\frac{\partial L}{\partial \mathbf{w}_{1}}=\mathbf{X}_1^T(\mathbf{X}_{1}\mathbf{w}_{1}+\mathbf{e}_{1}b_{1})+\mathbf{X}_2^T\mathbf{\upalpha}=0,
\end{aligned}
\end{equation}

\begin{equation}
\label{eq.difB}
\begin{aligned}
&\frac{\partial L}{\partial b_{1}}=\mathbf{e}_1^T(\mathbf{X}_{1}\mathbf{w}_{1}+\mathbf{e}_{1}b_{1})+\mathbf{e}_2^T\mathbf{\upalpha}=0,
\end{aligned}
\end{equation}

\begin{equation}
\label{eq.difX}
\begin{aligned}
&\frac{\partial L}{\partial \mathbf{\upxi}}=c_{1}\mathbf{\upxi}-\mathbf{\upalpha}=0,
\end{aligned}
\end{equation}

\begin{equation}
\label{eq.difAlpha}
\begin{aligned}
&\frac{\partial L}{\partial \mathbf{\upalpha}}=-(\mathbf{X}_{2}\mathbf{w}_{1}+\mathbf{e}_{2}b_{1})+\mathbf{\upxi}-\mathbf{e}_{2}=0.
\end{aligned}
\end{equation}

\noindent By combining two equations (\ref{eq.difW}) and (\ref{eq.difB}) we give the equation (\ref{eq.t1}) as follows:

\begin{equation}
\label{eq.t1}
\begin{bmatrix} \mathbf{X}_1^T\\\mathbf{e}_1^T \end{bmatrix}\begin{bmatrix} \mathbf{X}_1  &\mathbf{e}_1 \end{bmatrix}\begin{bmatrix} \mathbf{w}_1\\b_1\end{bmatrix}+\begin{bmatrix} \mathbf{X}_2^T\\\mathbf{e}_2^T \end{bmatrix}\mathbf{\upalpha}=0.
\end{equation}

\noindent Using equations (\ref{eq.difX}), (\ref{eq.difAlpha}) and (\ref{eq.t1}) the following values are obtained:

\begin{equation}
\label{eq.11}
\begin{bmatrix} \mathbf{w}_1\\b_1\end{bmatrix}=-(\mathbf{G}^T\mathbf{G}+\frac{1}{c_{1}}\mathbf{H}^{T}\mathbf{H})^{-1}\mathbf{G}^{T}\mathbf{e}_{2},
\end{equation}

\noindent where $\mathbf{H}=\begin{bmatrix} \mathbf{X}_1&\mathbf{e}_1\end{bmatrix}$ and $\mathbf{G}=\begin{bmatrix} \mathbf{X}_2&\mathbf{e}_2\end{bmatrix}$. In a similar manner, the parameters of the other hyper-plane are obtained as:

\begin{equation}
\label{eq.12}
\begin{aligned}
&\begin{bmatrix} \mathbf{w}_2\\b_2\end{bmatrix}=(\mathbf{H}^T\mathbf{H}+\frac{1}{c_{2}}\mathbf{G}^{T}\mathbf{G})^{-1}\mathbf{H}^{T}\mathbf{e}_{1}.
\end{aligned}
\end{equation}

After obtaining weights and biases of the two non-parallel hyper-planes using equations (\ref{eq.11}) and (\ref{eq.12}), we assign a new data point $\mathbf{x}$ into the class $i$ as follows:

\begin{equation}
\label{eq.classi}
i=arg\min_{j=1,2}\frac{\mid \mathbf{w}_j^T \mathbf{x}+b_{j}\mid}{\parallel \mathbf{w}_{j}\parallel}.
\end{equation}

\section{Rough Set theory}
\label{4RoughSetTheory}
Rough set theory is invented by Pawlak \cite{Pawlak1992}. This theory is concerned with analysis of deterministic data dependencies. It is a nondeterminism mathematical powerful tool for inference in the presence of ambiguity cases, that provides methods for removing irrelevant information from data. The indiscernibility concept is the basis of this theory. Consider an information system $\mathcal{I}=(\mathcal{U},\mathcal{S})$, where $\mathcal{U}$ is a set of finite instances and $\mathcal{S}$ is a set of their attributes, such that each attribute $a\in \mathcal{S}$ defines a mapping $a:\mathcal{U}\rightarrow \mathcal{V}_{a}$, where $\mathcal{V}_{a}$ is the set of values that attribute $a$ can take. For every $\mathcal{P}\subseteq \mathcal{S}$, the indiscernibility relation is:

\begin{equation}
\mathcal{R}_{\mathcal{P}}=IND(\mathcal{P})=\left\{(\mathbf{x},\mathbf{y}) \in \mathcal{U}^{2}\mid \forall a\in \mathcal{P}, a(\mathbf{x})=a(\mathbf{y}) \right\}.
\end{equation}

\noindent If $(\mathbf{x},\mathbf{y})\in IND(\mathcal{P})$, then $\mathbf{x}$ and $\mathbf{y}$ are indiscernible by attributes of $\mathcal{P}$. So, $R_{\mathcal{P}}$ partitions the set $\mathcal{U}$, and the equivalence classes of $\mathcal{P}$-indiscernibility relation are denoted by $[\mathbf{x}]_{\mathcal{P}}$. \par

Suppose $\mathcal{X}\subseteq \mathcal{U}$ and $\mathcal{B}\subseteq \mathcal{S}$ such that the relation $IND(\mathcal{\mathcal{B}})$ constructs a partitioning and creates equivalence classes. Accordingly, $\mathcal{X}$ can be approximated based on the information in the feature space $\mathcal{B}$ that is obtained by the lower bound of $\mathcal{X}$($\mathcal{B}\downarrow \mathcal{X}$), and the upper bound of $\mathcal{X}$ ($\mathcal{B}\uparrow \mathcal{X}$) is defined as follow:

\begin{equation}
\mathcal{B}\downarrow \mathcal{X}=\left\{\mathbf{x}\mid [\mathbf{x}]_{\mathcal{B}}\subseteq \mathcal{X}\right\},
\end{equation}

\begin{equation}
\mathcal{B}\uparrow \mathcal{X}=\left\{\mathbf{x}\mid [\mathbf{x}]_{\mathcal{B}}\cap \mathcal{X} \neq \varnothing\right\}.
\end{equation}

The boundary region can be defined based on the upper and lower bound as follows:

\begin{equation}
BN_{\mathcal{B}}(\mathbf{x})=\mathcal{B}\uparrow \mathcal{X}-\mathcal{B}\downarrow \mathcal{X}.
\end{equation}

\noindent If $BN_{\mathcal{B}}(\mathbf{x})\neq\varnothing$ then $\mathcal{X}$ is a rough set. So, a rough set is recognized as $\langle \mathcal{B}\uparrow \mathcal{X}, \mathcal{B}\downarrow \mathcal{X}\rangle$.

A special kind of an information system (IS) is used in classification or prediction tasks, which is called decision system $(\mathcal{U},\mathcal{S}\cup\left\{d\right\})$, where $d$ is a decision attribute. In case $d$ is nominal, the equivalence classes $[\mathbf{x}]_{R_{d}}$, resulted from indiscernibility relation $R$ on attribute $d$, are called decision classes.\par

For every $\mathcal{B}\subseteq \mathcal{S}$ in $IS$, the $\mathcal{B}$-positive region is defined as the set $\mathcal{X}\subseteq \mathcal{U}$ which is identically discernible for values of $\mathcal{B}$, as follow:

\begin{equation}
POS_{\mathcal{B}}=\cup_{\mathbf{x}\in \mathcal{X}}R_{\mathcal{B}}\downarrow[\mathbf{x}]_{R_{d}}.
\end{equation}

\noindent Two examples $\mathbf{x}$ and $\mathbf{y}$ in $POS_{\mathcal{B}}$ have equivalent values for attributes in $\mathcal{B}$, which means that those examples belong to the same class.

\subsection{Fuzzy-Rough Set}
\label{4.1Fuzzy-RoughSet}
In practice, datasets contain attributes with exact and real values. In these cases, rough set theory is confronted with problem. In this theory, we cannot specify that the values of two attributes are similar or not, or specify the amount of similarity between them. For example, two numbers $-0.123$ and $-0.122$ may be different only because of noise, but rough set theory considers those numbers different. One way to cope with this problem is discretization of the database. But, this method loses information. Fuzzy-rough theory is the basic solution in these cases. In the early 1990s, researches on combining fuzzy sets and rough sets are emerged \cite{432167}. According to the following guiding principles, fuzzy rough set theory has focused mainly on fuzzifying the lower and upper approximations:

\begin{itemize}
\item In order to generalize the set $\mathcal{X}$ to a fuzzy set in $\mathcal{U}$, it allows objects to belong to a given concept with various degrees of membership.
\item Unlike rough set theory that evaluates objects by indiscernibility relation, a fuzzy relation $R$ is used to approximate equality of objects. As a result, objects can be categorized into classes based on their similarity to each other with soft boundaries and objects can belong to more than one class with varying degrees.
\end{itemize}

More formally, equality of objects is modeled with fuzzy relation $R$ in $\mathcal{X}$, such that we can assign a similarity degree to each pair of objects.\par

The classical way of objects discerning is used for objects with the qualitative attribute $a$, i.e., $R_{a} (\mathbf{x},\mathbf{y})=1$ if $a(\mathbf{x})=a(\mathbf{y})$ and $R_{a}(\mathbf{x},\mathbf{y})=0$ otherwise. The Fuzzy $\mathcal{B}$-indiscernibility relation for any subset $\mathcal{B}$ of $\mathcal{S}$ is defined in equation (\ref{eg.fuzzyBIndis}) (there are several definitions in fuzzy relation $R$ and fuzzy set $\mathcal{A}$ for upper and lower approximations of $\mathcal{A}$ by $R$):

\begin{equation}
\label{eg.fuzzyBIndis}
R_{\mathcal{B}}(\mathbf{x},\mathbf{y})=\mathcal{T}(\underbrace{R_{a}(\mathbf{x},\mathbf{y}))}_{a\in \mathcal{B}},
\end{equation}

\noindent where $\mathcal{T}(\cdot)$ represents a fuzzy $t$-norm. It can be seen that by using only qualitative attributes, the above relation is equivalent to the traditional concept of the $\mathcal{B}$-indiscernibility relation.\par

By using fuzzy relation $R$, the lower and upper bound of a fuzzy set $\mathcal{A}$ in $\mathcal{X}$ are defined in the following forms:

\begin{equation}
(R_{b}\downarrow \mathcal{A})(\mathbf{x})=\inf_{\mathbf{y} \in \mathcal{X}} \mathcal{L}(R(\mathbf{x},\mathbf{y}),\mathcal{A}(\mathbf{y})),
\end{equation}

\begin{equation}
(R_{b}\uparrow \mathcal{A})(\mathbf{x})=\sup_{\mathbf{y} \in \mathcal{X}} \mathcal{T}(R(\mathbf{x},\mathbf{y}),\mathcal{A}(\mathbf{y})),
\end{equation}

\noindent where $\mathcal{L}(\cdot,\cdot)$ is a fuzzy implication and $\mathcal{T}(\cdot)$ is a fuzzy $t$-norm. Each mapping in the the form $[0,1]^2\rightarrow [0,1]$ such that for each $x\in [0,1]$ satisfy $L(0,0)=1$ and $L(1,x)=x$ is a fuzzy implication. \par

The fuzzy $\mathcal{B}$-positive region for $\mathbf{y}$ in $\mathcal{X}\subseteq U$ can be defined as:

\begin{equation}
POS_{\mathcal{B}}(\mathbf{y})=(\cup_{\mathbf{x}\in \mathcal{X}}R_{\mathcal{B}}\downarrow R_{d}\mathbf{x})(\mathbf{y}),
\end{equation}

\noindent where $d$ is a decision attribute.

\section{Proposed method}
\label{5ProposedMethod}
The fuzzy rough weighted least squares twin support vector machine (FRLSTSVM) for imbalanced data classification is presented in this section. The proposed FRLSTSVM method reduces the training points in the majority class using the fuzzy-rough lower approximation and determines the weight biases in the formulation of WLTSVM using fuzzy rough set theory.

\subsection{Linear FRLSTSVM}
\label{5.1LinearFRLSTSVM}
In order to generate two proximal hyperplanes with WLTSVM, we suppose that data points in the minority class remain unchanged, while data points in the majority class are reduced with the method that we present its formulation in the following.\par

First, the $\mathcal{B}$-positive region for majority data points is calculated, where $\mathcal{B}$ is the set of all attributes. Obviously, data points in the high-density regions are similar to each other, and therefore these data points have a high fuzzy $\mathcal{B}$-positive degree, while data points in the low-density regions have a low $\mathcal{B}$-positive degree, e.g. outliers data. Finally, for each majority data point $\mathbf{x}_{i}$, if the fuzzy $\mathcal{B}$-positive degree of $\mathbf{x}_{i}$ is more than pre-defined non-negative threshold $\tau$, then $\mathbf{x}_{i}$ can be selected in the final majority dataset.\par

Given a decision system $(\mathcal{U},\mathcal{S}\cup\left\{d\right\})$, where $\mathcal{U}$ is a non-empty set of majority data points, $\mathcal{S}$ is a non-empty finite set of the attributes and $d$ is a decision attribute. By considering $a$ as a quantitative attribute in $\mathcal{S}\cup\left\{d\right\}$, we can define the approximate equality between two data points $\mathbf{x}$ and $\mathbf{y}$ in $a$ with the following fuzzy relation:

\begin{equation}
\label{eq23}
R_a^\gamma=max(0,1-\gamma\frac{\mid a(\mathbf{x})-a(\mathbf{y})\mid}{l(a)}),
\end{equation}

\noindent where parameter $\gamma>0$ determines the granularity of $R_a^\gamma$ and $l(a)$ is the range of the attribute $a$. In order to determine the similarity between $\mathbf{x}$ and $\mathbf{y}$ based on any subset $\mathcal{B}$ of $\mathcal{S}$, fuzzy $\mathcal{B}$-indiscernibility can be defined as follows:

\begin{equation}
\label{eg.fuzzyBIndisRel}
R^\gamma_{\mathcal{B}}(\mathbf{x},\mathbf{y})=\mathcal{T}(\underbrace{R^\gamma_{a}(\mathbf{x},\mathbf{y}))}_{a\in \mathcal{B}}.
\end{equation}

The lower approximation $R^\gamma_{\mathcal{B}}\downarrow \mathcal{A}$ from the fuzzy set $\mathcal{A}$ in $\mathcal{S}$ by using the fuzzy relation $R^\gamma_{\mathcal{B}}$ for $\mathbf{y}\in \mathcal{X}$ can be defined as:

\begin{equation}
(R^\gamma_{b}\downarrow \mathcal{A})(\mathbf{y})=\inf_{\mathbf{x} \in \mathcal{X}} \mathcal{L}(R^\gamma_{\mathcal{B}}(\mathbf{x},\mathbf{y}),\mathcal{A}(\mathbf{x})).
\end{equation}

\noindent Thus, we can define the the fuzzy $\mathcal{B}$-positive region $POS^{\gamma,\mathcal{X}}_{\mathcal{B}}$ for $\mathbf{y}\in \mathcal{X}$ as:

\begin{equation}
\label{eq.POS}
POS^{\gamma,\mathcal{X}}_{d}(\mathbf{y})=(R^\gamma_{\mathcal{B}}\downarrow R^\gamma_{d})(\mathbf{y}).
\end{equation}

In order to reduce the majority class data points, in the first step, the fuzzy $\mathcal{B}$-indiscernibility relations for all data points in the the majority class are calculated, where $\mathcal{B}$ is the set of all attributes approximating equality of objects with their similar ones. Since we use this method only to reduce data points with decision attribute $d=-1$ (majority class), the fuzzy $\mathcal{B}$-indiscernibility relation for all data points is equivalent to the fuzzy $\mathcal{B}$-positive region, and this algorithm assigns the membership degree of each training data in positive region.\par

In the second step, by using a threshold value $\tau$, if the similarity of data points (appertain to $POS^{\gamma,\mathcal{X}}_{d}$) is less than $\tau$, then these data points are finally not selected for reduced dataset.\par

Since the membership degrees of the lower fuzzy-rough approximation in the boundary or overlapped regions are less than the membership degrees of data points located in the centers regions, it can be concluded that the lower fuzzy-rough approximation is suitable for verification of the quality of neighbors for any data point.\par

This fuzzy-rough set sub-sampling method requires the input parameters $\gamma$ as the fuzzy similarity relation and $\tau$ which is used as a threshold to remove the objects having fewer values than this threshold. \par

In the proposed approach, we design a suitable selection weight approach for constructing two proximal hyperplanes. Weights should fulfil two conditions: 1) a higher weight value should be assigned to the minority class data points, while the majority class data points should receive a lower weight, and 2) the weights should be in the interval $[0,1]$, so that the training of the classifier can converge. Also, as the second contribution, we propose an approach to assign weight to the majority and minority data points. For each data point in the majority class, first, its fuzzy $\mathcal{B}$-indiscernibility relation with all other majority calss data points is calculated, where $\mathcal{B}$ is the set of all attributes that the fuzzy $\mathcal{B}$-indiscernibility relation with decision attribute $d=-1$ is equivalent to the fuzzy $\mathcal{B}$-positive region. Then, the weight of each majority class data point is equivalent to the average membership degree of its related points in the fuzzy $\mathcal{B}$-positive region. For each data point $\mathbf{x}_{i}$ in the majority class, the corresponding weights $v_{i}$ calculates using the following equation:

\begin{equation}
\label{eq.wieghtMaj}
\forall \mathbf{x}_{i}\in \mathbf{X}_{2}:   v_{i}=\frac{1}{n}\sum_{\mathbf{y}=\mathbf{x}^{1}_{2}, \mathbf{y}\neq \mathbf{x}^{i}_{2} }^{\mathbf{x}^{n}_{2}} POS^{\gamma, \mathbf{y}}_{d=-1}(\mathbf{x}_{i}),
\end{equation}

\noindent where $n$ is the number of data points in the majority class. We define the diagonal weight matrix $\mathbf{D}_{2} (n_2\times n_2)$ having weights $v_{i}$ which corresponded to the majority class data points as diagonal elements.\par

Also, for each minority class data point, first, its fuzzy $\mathcal{B}$-indiscernibility relation of all majority class data points is calculated. Then, the weight of each minority data point is equivalent to the average membership degree of its related points in the fuzzy $\mathcal{B}$-positive region. For each data point $\mathbf{x}_{j}$ in the minority class, the corresponding weights $v_{j}$ calculates using the following equation:

\begin{equation}
\label{eq.wieghtMin}
\forall \mathbf{x}_{j}\in \mathbf{X}_{1}:   v_{j}=\frac{1}{n}\sum_{\mathbf{y}=\mathbf{x}^{1}_{1}, \mathbf{y}\neq \mathbf{x}^{j}_{1} }^{\mathbf{x}^{n}_{1}} POS^{\gamma, \mathbf{y}}_{d}(\mathbf{x}_{j}).
\end{equation}

\noindent We define the diagonal weight matrix $\mathbf{D}_{1}(n_1\times n_1)$ having weights $v_{j}$, which its diagonal elements correspond to the minority class data points.\par

By using $\mathbf{X}_{1}$, $\mathbf{\widehat{X}}_{2}$, $\mathbf{D}_{1}$, and $\mathbf{D}_{2}$, which $\mathbf{\widehat{X}}_{2}$ is a reduced matrix of majority data points, we determine two hyperplanes by solving the following optimization problems:

\begin{equation}
\begin{split}
\begin{aligned}
&\min_{\mathbf{w}_{1},b_{1},\mathbf{\upxi}} \frac{1}{2}\parallel \mathbf{X}_{1}\mathbf{w}_{1}+\mathbf{e}_{1}b_{1}\parallel^{2}+\frac{c_{1}}{2}\mathbf{\upxi}^{T}\sum_{i=1}^{m_2}v_{i}\mathbf{\upxi},
\\
&s.t. \quad -(\mathbf{\widehat{X}}_{2}\mathbf{w}_{1}+\mathbf{e}_{2}b_{1})+\mathbf{\upxi}\geq \mathbf{e}_{2}, \quad \mathbf{\upxi}\geq0
\end{aligned}
\end{split}
\end{equation}

\noindent and

\begin{equation}
\begin{split}
\begin{aligned}
&\min_{\mathbf{w}_{2},b_{2},\mathbf{\upeta}} \frac{1}{2}\parallel \mathbf{\widehat{X}}_{2}\mathbf{w}_{2}+\mathbf{e}_{2}b_{2}\parallel^{2}+\frac{c_{2}}{2}\mathbf{\upeta}^{T}\sum_{i=1}^{m_1}v_{j} \mathbf{\upeta},
\\
&s.t. \quad (\mathbf{X}_{1}\mathbf{w}_{2}+\mathbf{e}_{1}b_{2})+\mathbf{\upeta}\geq \mathbf{e}_{1}, \quad \mathbf{\upeta}\geq0
\end{aligned}
\end{split}
\end{equation}

\noindent where $\mathbf{\upxi}$ and $\mathbf{\upeta}$ are slack variables and $\mathbf{e}_{1}$ and $\mathbf{e}_{2}$ are two vectors having suitable dimensions with all values as $1$s. The parameters $c_{1}>0$ and $c_{2}>0$ are penalty factors and $m_{1}$ and $m_{2}$ are number of data points in the minority and majority classes respectively. Lagrangians of the above equations are given as:

\begin{equation}
\label{eq.lagra}
\begin{aligned}
L(\mathbf{w}_{1},b_{1},\mathbf{\upxi},\mathbf{\upalpha})&=\frac{1}{2}\parallel \mathbf{X}_{1}\mathbf{w}_{1}+\mathbf{e}_{1}b_{1}\parallel^{2}+\frac{c_{1}}{2}\mathbf{\upxi}^{T}v_{i}\mathbf{\upxi}\\
&-\mathbf{\upalpha}^{T}(-(\mathbf{\widehat{X}}_{2}\mathbf{w}_{1}+\mathbf{e}_{2}b_{1})+\mathbf{\upxi}-\mathbf{e}_{2}),
\end{aligned}
\end{equation}

\noindent and

\begin{equation}
\begin{aligned}
L(\mathbf{w}_{2},b_{2},\mathbf{\upeta},\mathbf{\upbeta})&=\frac{1}{2}\parallel \mathbf{\widehat{X}}_{2}\mathbf{w}_{2}+\mathbf{e}_{2}b_{2}\parallel^{2}+\frac{c_{2}}{2}\mathbf{\upeta}^{T}v_{j}\mathbf{\upeta}\\
&-\mathbf{\upbeta}^{T}((\mathbf{X}_{1}\mathbf{w}_{2}+\mathbf{e}_{1}b_{2})+\mathbf{\upeta}-\mathbf{e}_{1}),
\end{aligned}
\end{equation}

\noindent where $\mathbf{\upalpha}\in \mathbb{R}^{n}$ and $\mathbf{\upbeta}\in \mathbb{R}^{n}$ are two Lagrangian coefficient vectors. According to KKT conditions, the corresponding derivatives of equation (\ref{eq.lagra}) are obtained and set to zero. As a result, equations (\ref{eq.difW}), (\ref{eq.difB}) and (\ref{eq.difAlpha}) are again obtained, and following equation is obtained instead of equation (\ref{eq.difX}):

\begin{equation}
\label{eq.diffXi}
\frac{\partial L}{\partial \mathbf{\upxi}}=c_{1}v_{i}\mathbf{\upxi}-\mathbf{\upalpha}=0.
\end{equation}

\noindent Considering two matrices $\mathbf{H}=\begin{bmatrix}\mathbf{X}_1&\mathbf{e}_1\end{bmatrix}$ and $\mathbf{G}=\begin{bmatrix}\mathbf{\widehat{X}}_2&\mathbf{e}_2\end{bmatrix}$ and two vectors $\mathbf{u}_{1}=\begin{bmatrix}\mathbf{w}_1 \\ b_1\end{bmatrix}$ and $\mathbf{u}_{2}=\begin{bmatrix}\mathbf{w}_2 \\ b_2\end{bmatrix}$, equation (\ref{eq.diffXi}) can be rewritten as:

\begin{equation}
\label{eq.40}
\mathbf{H}^T\mathbf{H}\mathbf{u}_{1}+\mathbf{G}^T \mathbf{\upalpha}=0.
\end{equation}

\noindent After solving this equation, we find:

\begin{equation}
\label{eq35}
\mathbf{u}_{1}=-(\mathbf{H}^T \mathbf{H})^{-1}\mathbf{G}^T \mathbf{\upalpha}.
\end{equation}

\noindent In order to avoid the possible ill-conditioning of $\mathbf{H}^T \mathbf{H}$ in the special case, which it may be singular, TSVM introduces a term $\delta \mathbf{I} (\delta>0)$, and replaces $\mathbf{H}^T \mathbf{H}$ by $\mathbf{H}^T \mathbf{H}+\delta \mathbf{I}$, where $\mathbf{I}$ is an identity matrix of appropriate dimensions. In this case, equation (\ref{eq35}) can be rewritten as follows:

\begin{equation}
\label{eq.u1}
\mathbf{u}_{1}=-(\mathbf{H}^T \mathbf{H}+\delta \mathbf{I})^{-1}\mathbf{G}^T \mathbf{\upalpha}.
\end{equation}

\noindent After solving equations (\ref{eq.difAlpha}), (\ref{eq.diffXi}) and (\ref{eq.40}) we obtain:

\begin{equation}
\label{eq.alpha}
\mathbf{\upalpha}=[\frac{\mathbf{D}_i^{-1}}{c_{1}}+\mathbf{G}(\mathbf{H}^T\mathbf{H})^{-1}\mathbf{G}^T]^{-1}\mathbf{e}_{2},
\end{equation}

\noindent where $\mathbf{D}_{i}$ is a diagonal matrix having weights $v_{i}$ which corresponded to the majority class data points. \par

\noindent In the same way:

\begin{equation}
\label{eq.u2}
\mathbf{u}_{2}=(\mathbf{G}^T \mathbf{G}+\delta \mathbf{I})^{-1}\mathbf{H}^T \mathbf{\upbeta},
\end{equation}
and
\begin{equation}
\label{eq.beta}
\mathbf{\upbeta}=[\frac{\mathbf{D}_j^{-1}}{c_{2}}+\mathbf{H}(\mathbf{G}^T \mathbf{G})^{-1}\mathbf{H}^T]^{-1}\mathbf{e}_{1},
\end{equation}

\noindent where $\mathbf{D}_{j}$ is a diagonal matrix having weights $v_{j}$ which corresponded to the minority class data points.\par

Once solutions $\mathbf{w}_{1}$, $b_{1}$, $\mathbf{w}_{2}$ and $b_{2}$ are obtained, for each new data point $\mathbf{x}\in \mathbb{R}^n$, first its distance from the hyperplanes are calculated, then it  assigns to the class $i$ $(i=+1,-1)$ having the smallest distance as follows:

\begin{equation}
\label{eq.newClassify}
i=arg\min_{j=1,2}\frac{\mid \mathbf{w}_j^Tx+b_{j}\mid}{\parallel \mathbf{w}_{j}\parallel}.
\end{equation}

\subsection{Nonlinear FRLSTSVM}
\label{5.2NonlinearFRLSTSVM}

To extend the proposed model to the nonlinear case, we consider two following kernel based surfaces instead of two hyperplanes:

\begin{equation}
\label{eqNon1}
\begin{split}
\begin{aligned}
&f_1(\mathbf{x})=K(\mathbf{X}_{1},\mathbf{X})\mathbf{w}_{1}+b_{1}=0,\\
&f_2(\mathbf{x})=K(\mathbf{X}_{2},\mathbf{X})\mathbf{w}_{2}+b_{2}=0,
\end{aligned}
\end{split}
\end{equation}

\noindent where $K$ is an appropriately chosen kernel function and $\mathbf{X}=[\mathbf{X}_{1},\mathbf{X}_{2}]^T$. After solving two following kernel based optimization problems, FRLSTSVM for nonlinear data classification will be obtained as follows:

\begin{equation}
\label{eqNon2}
\begin{split}
\begin{aligned}
&\min_{\mathbf{w}_{1},b_{1},\mathbf{\upxi}} \frac{1}{2}\parallel K(\mathbf{X}_{1},\mathbf{X})\mathbf{w}_{1}+\mathbf{e}_{1}b_{1}\parallel^{2}+\frac{c_{1}}{2}\mathbf{\upxi}^{T}\sum_{i=1}^{m_2}v_{i}\mathbf{\upxi},
\\
&s.t. \quad -(K(\mathbf{\widehat{X}}_{2},\mathbf{X})\mathbf{w}_{1}+\mathbf{e}_{2}b_{1})+\mathbf{\upxi}\geq \mathbf{e}_{2}, \quad \mathbf{\upxi}\geq0
\end{aligned}
\end{split}
\end{equation}

\noindent and \par

\begin{equation}
\label{eqNon3}
\begin{split}
\begin{aligned}
&\min_{\mathbf{w}_{2},b_{2},\mathbf{\upeta}} \frac{1}{2}\parallel K(\mathbf{\widehat{X}}_{2},\mathbf{X})\mathbf{w}_{2}+\mathbf{e}_{2}b_{2}\parallel^{2}+\frac{c_{2}}{2}\mathbf{\upeta}^{T}\sum_{i=1}^{m_1}v_{j}\mathbf{\upeta},
\\
&s.t. \quad (K(\mathbf{X}_{1},\mathbf{X})\mathbf{w}_{2}+\mathbf{e}_{1}b_{2})+\mathbf{\upeta}\geq \mathbf{e}_{1}, \quad \mathbf{\upeta}\geq0
\end{aligned}
\end{split}
\end{equation}

\noindent where $\mathbf{X}=[\mathbf{\widehat{X}}_{1},\mathbf{X}_{2}]^T$. Considering two matrices  $\mathbf{P}=\begin{bmatrix}K(\mathbf{X}_1,\mathbf{X})&\mathbf{e}_1\end{bmatrix}$ and $\mathbf{Q}=\begin{bmatrix}K(\mathbf{\widehat{X}}_2,\mathbf{X})&\mathbf{e}_2\end{bmatrix}$ and two vectors $\mathbf{z}_{1}=\begin{bmatrix}\mathbf{w}_1\\b_1\end{bmatrix}$ and $\mathbf{z}_{2}=\begin{bmatrix}\mathbf{w}_2\\b_2\end{bmatrix}$, the solution of the above optimization problems will be found as:

\begin{equation}
\label{eqNon4}
\mathbf{z}_{1}=-(\mathbf{P}^T \mathbf{P}+\delta \mathbf{I})^{-1}\mathbf{Q}^T \mathbf{\upalpha},
\end{equation}

\begin{equation}
\label{eqNon5}
\mathbf{\upalpha}=[\frac{\mathbf{D}_i^{-1}}{c_{1}}+\mathbf{Q}(\mathbf{P}^T\mathbf{P})^{-1}\mathbf{Q}^T]^{-1}\mathbf{e}_{2},
\end{equation}

\begin{equation}
\label{eqNon6}
\mathbf{z}_{2}=(\mathbf{Q}^T \mathbf{Q}+\delta \mathbf{I})^{-1}\mathbf{P}^T \mathbf{\upbeta},
\end{equation}

\begin{equation}
\label{eqNon7}
\mathbf{\upbeta}=[\frac{\mathbf{D}_j^{-1}}{c_{2}}+\mathbf{P}(\mathbf{Q}^T\mathbf{Q})^{-1}\mathbf{P}^T]^{-1}\mathbf{e}_{1},
\end{equation}

\noindent which are the solutions for $\mathbf{w}_{1}$, $b_{1}$, $\mathbf{w}_{2}$ and $b_{2}$. \par

A new data point $\mathbf{x}\in \mathbb{R}^{n}$ is then assigned to the class $i$ $(i=+1,-1)$ using the following decision rule:

\begin{equation}
\label{eqNon8}
i=arg\min_{j=1,2}\frac{\mid K(\mathbf{x},\mathbf{X})\mathbf{w}_{j}+b_{j}\mid}{\sqrt{K(\mathbf{x},\mathbf{X})\mathbf{w}_{j}+b_{j}}}
\end{equation}

Summary of the proposed method is shown in algorithm (\ref{alg1}).

%Algorithm 1 ****************

\begin{algorithm}[t]
\caption{FRLSTSVM ($\mathcal{X}_{+}$, $\mathcal{X}_{-}$, $\mathcal{S}$, $d$, $\tau, \gamma$)}
\label{alg1}
 \textbf{Input:} \\
 $\mathcal{X}_{+}$, the set of positive class instances.\\
 $\mathcal{X}_{-}$, the set of negative class instances.\\
 $\mathcal{S}$, non-empty finite set of attributes.\\
 $d$, a decision attribute belong to $\left\{1,-1\right\}$.\\
\textbf{Hyper-parameters:} \\
 $\gamma$, granularity parameter.\\
 $\tau$, selection threshold.\\
 \textbf{Output:}\\
 $\mathbf{w}_{1}$, $b_{1}$, $\mathbf{w}_{2}$, and $b_{2}$.\\
  \textbf{Process:}\\
  $\mathbf{X}_{1}\leftarrow \mathcal{X}_{+}$\\
  $\mathbf{\widehat{X}}_{2}\leftarrow \mathcal{X}_{-}$\\
A) subsampling of majority data:
  \begin{enumerate}

\item \textbf{For} any subset $\mathcal{B}$ of $\mathcal{S}$, Calculate fuzzy $\mathcal{B}$-positive region $POS_\mathcal{B}^{\gamma,\mathcal{X}_{-}}$ with equation (\ref{eq.POS}).
\item \textbf{For each} $\mathbf{x} \in \mathcal{X}_{-}$:\\
\hspace*{1cm}\textbf{If} $(POS_B^{\gamma,\mathcal{X}_{-}}(\mathbf{x})<\tau)$\\
\hspace*{2cm}	$\mathbf{\widehat{X}}_{2}\leftarrow \mathbf{\widehat{X}}_{2}-\left\{\mathbf{x}\right\}.$

\end{enumerate}
B) assigns weights to the majority and minority data points using equations (\ref{eq.wieghtMaj}) and (\ref{eq.wieghtMin}) respectively.\\
C) calculate hyperplanes:
  \begin{enumerate}
\item calculate $\mathbf{\upalpha}$ and $\mathbf{\upbeta}$ using equations (\ref{eq.alpha}) and (\ref{eq.beta}) respectively.
\item calculate $\mathbf{u}_{1}$ and $\mathbf{u}_{2}$ using equations (\ref{eq.u1}) and (\ref{eq.u2}) respectively.
\item calculate $\mathbf{w}_{1}$, $b_{1}$, $\mathbf{w}_{2}$, and $b_{2}$: $\mathbf{u}_{1}=\begin{bmatrix}\mathbf{w}_{1} &b_{1}\end{bmatrix} ^{T}$ and $\mathbf{u}_{2}=\begin{bmatrix}\mathbf{w}_{2} &b_{2}\end{bmatrix} ^{T}$.
\end{enumerate}
\textbf{Return:} $\mathbf{w}_{1}$, $b_{1}$, $\mathbf{w}_{2}$, and $b_{2}$.

D) Classification of new sample using equation (\ref{eq.newClassify}).
\end{algorithm}

\section{Experimental results}
\label{6ExperimentalResults}
In order to evaluate the proposed FRLSTSVM method, $15$ independent imbalanced datasets are utilized to investigate its performance. Table (\ref{Table.Param}) shows the details of these datasets that constitute several imbalanced classification problems of various sizes, imbalance ratio and number of features. In each dataset, the minority class is marked and the rest classes are majority class. The seven datasets Yeast3, Vehicle, PimaIndian, Haberman, Yeast4, Shuttle, Segment are imbalanced datasets from Knowledge Extraction based on the Evolutionary Learning (KEEL\footnote{https://sci2s.ugr.es/keel/imbalanced.php}), and the other datasets are from the UCI Repository\footnote{https://archive.ics.uci.edu/ml/index.php} that have not generally been assigned as imbalanced datasets.

\begin{table}[ht]
\caption {Characteristics of the benchmark two-class datasets}
\label{Table.Param}
\begin{tabular}{ccccc}
\hline
\multicolumn{1}{c}{\multirow{2}{*}{Dataset}} & \multicolumn{1}{c}{Imbalance} & \multirow{2}{*}{Features} & \multicolumn{1}{c}{Minority} & \multicolumn{1}{c}{Data}    \\
\multicolumn{1}{c}{}                         & \multicolumn{1}{c}{Ratio}     &                           & \multicolumn{1}{c}{Class}    & \multicolumn{1}{c}{Samples} \\ \hline
   Yeast3 & 8.1075 & 8 & ME3 & 1484 \\
    Vehicle & 3.2337 & 18 & VAN & 940 \\
    Transfusion & 3.2017 & 4 & Yes & 748 \\
    Wine & 2.0166 & 13 & Class 1 & 178 \\
    PimaIndian & 1.8653 & 8 & Positive & 768 \\
    Ionosphere & 1.7855 & 34 & Bad & 351 \\
    Haberman & 2.7779 & 3 & Died & 306 \\
    CMC & 3.4228 & 9 & Long-term & 1473 \\
    Vowel & 9.1010 & 13 & Class0 & 988 \\
    Yeast4 & 28.4118 & 8 & ME4 & 1484 \\
    shuttle-c0-vs-c4 & 13.8810 & 9 & Class0 & 1829 \\
    Segment & 6.0077 & 19 & Segment & 2308 \\
    Abalone19 & 128.8701 & 8 & Class19 & 4174 \\
    Wisconsin & 8.7943 & 9 & Rest & 683 \\
    Led7digit & 8.7943 & 7 & Rest & 443 \\
    \hline
\end{tabular}
\end{table}

In the experiments, in order to demonstrate the superiority of the proposed FRLSTSVM method, we compare it to the several SVM-based methods such as WSVM \cite{Fan2008}, TWSVM \cite{5762620,4135685}, SMOTE under-sampling \cite{DBLP2011}, SVM with under-sampling and over-sampling method \cite{4717268,Wang2010} and WLTSVM \cite{SHAO20143158}, using datasets shown in the Table (\ref{Table.Param}). The proposed FRLSTSVM method is implemented\footnote{The Matlab code of the FRLSTSVM method is available at https://github.com/maysambehmanesh/FRLSTSVM.} in MATLAB $R2017b$ on a PC with Intel $corei5$ processor $(2.4 GHz)$, $4.0$ GB of Ram and Windows $10$ OS.

%Most of the model parameters, such as the selection parameter $\tau$, the granularity parameter $\gamma$, and the parameters related to TSVM are found analytically, without using any complicated or time-consuming tuning procedure.

\subsection{Evaluation metrics}
\label{6.1EvaluationMeasures}
In the traditional binary classification tasks, the most important criterion is the accuracy, but this criterion is not appropriate for the imbalanced data classification. Accuracy, sensitivity, specificity, and G-mean, defined in equations (\ref{eq.Sen}) - (\ref{eq.Gmean}), are the classification criteria that compute from the confusion matrix shown in Table (\ref{Table.confMat}). G-mean as an appropriate criterion that can characterize the trade-off between sensitivity and specificity is widely used in the imbalanced data classification tasks.

\begin{table}
\caption {Confusion matrix}
\label{Table.confMat}
\begin{center}

	\begin{tabular}{ccc}
	\hline
	 & Predicted positive & Predicted negative\\
	\hline
	Actual positive & \textit{TP} & \textit{FN}\\
	Actual negative & \textit{FP} & \textit{TN}\\
		\hline
\end{tabular}
\end{center}
\end{table}

\begin{equation}
\label{eq.Sen}
Sen=TP/(TP+FP)
\end{equation}
\begin{equation}
\label{eq.Spe}
Spe= TN/(TN+FN)
\end{equation}
\begin{equation}
\label{eq.Acc}
Acc=(TP+TN)/(TP+FP+TN+FN)
\end{equation}
\begin{equation}
\label{eq.Gmean}
G-mean=\sqrt{Sen\times Spe}
\end{equation}

\subsection{Numerical results}
\label{6.2NumericalResults}
In this section, we evaluate the proposed FRLSTSVM method on the benchmark datasets and compare it to the most of the SVM-based state-of-the-art methods. In each dataset, we apply nested $10$-fold cross-validation technique. The whole process is repeated ten times and the mean accuracy (and G-mean) of these ten runs and theirs standard deviations are reported as the final results.\par

\begin{figure}[t]
	\centering
	\includegraphics[width=3.2in]{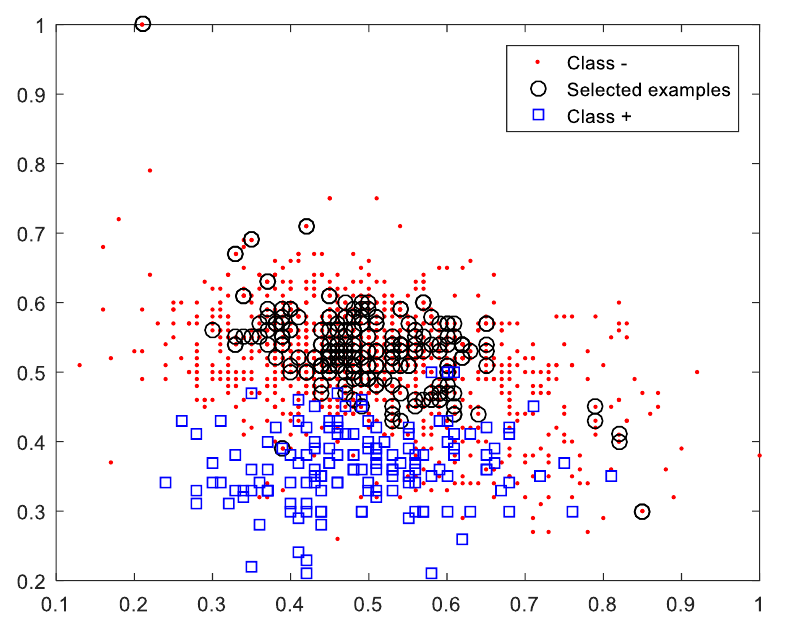}
	\caption{The selected majority data points of Yeast3 dataset after applying the proposed subsampling method. Small circles indicate the selected majority data points.}
	\label{fig.Yeast3}
\end{figure}

Figure (\ref{fig.Yeast3}) shows the selected majority data points of Yeast3 dataset after applying the proposed method. To be viewable, we only consider the first two features of the this dataset. This figure shows that the majority data points in the higher density regions have a good chances of selection in the final dataset, while the data points in the lower density regions (outlier examples), are unlikely to be selected in the final dataset.

The average accuracies of the FRLSTSVM method compared to the other methods with linear and non-linear kernels are summarized in Tables (\ref{Table.AccLin}) and (\ref{Table.accNlin}) respectively. Tables (\ref{Table.Gmlin}) and (\ref{Table.Gmnlin}) show the average G-mean of the FRLSTSVM and compare it to the other methods contain linear and non-linear kernels.

\begin{table*}[!]
\caption {The testing results of linear classifiers (Mean Accuracy($\%$)$\pm$ Standard Deviation)}
\label{Table.AccLin}
\begin{center}
\begin{tabular*}{\textwidth}{l l l l l l l l }

	\hline
	Dataset & WSVM & $Over_{SVM}$ & $Under_{SVM}$ & $SMOTE_{SVM}$ & TWSVM & WLTSVM & FRLSTSVM\\
	\hline
Yeast3 & $90.79\pm1.36$ & $90.22\pm0.15$ & $83.78\pm0.19$ & $91.92\pm0.39$ & $89.17\pm0.41$ & $91.96\pm0.19$ & \textbf{93.21}$\pm$\textbf{0.39}\\
Vehicle & $95.80\pm0.86$ & $94.67\pm0.10$ & $93.83\pm0.85$ & \textbf{97.03}$\pm$\textbf{0.24} & $96.59\pm0.23$ & $96.51\pm0.28$ & $96.21\pm0.31$\\
Transfusion & $62.73\pm4.28$ & $54.86\pm2.21$ & $55.69\pm2.67$ & $65.06\pm3.73$ & $49.43\pm2.27$ & $66.64\pm0.62$ & \textbf{72.45}$\pm$ \textbf{0.21}\\
Wine & $92.12\pm0.50$ & $95.72\pm0.55$ & $91.20\pm0.73$ & $96.04\pm0.19$ & $94.81\pm0.50$ & $95.12\pm0.8 7$ & \textbf{97.19}$\pm$ \textbf{0.11}\\
PimaIndian & $75.64\pm0.79$ & $75.26\pm0.90$ & $72.79\pm0.53$ & $73.01\pm0.62$ & $76.10\pm0.25$ & $76.78\pm0.35$ & \textbf{78.69}$\pm$ \textbf{0.45}\\
Ionosphere & \textbf{88.75}$\pm$\textbf{2.29} & $89.11\pm0.67$ & $83.62\pm0.44$ & $73.01\pm0.30$ & $84.93\pm1.21$ & $87.53\pm0.81$ & $85.48\pm0.26$\\
Haberman & $70.68\pm0.92$ & $64.82\pm0.51$ & $58.96\pm1.74$ & $74.77\pm0.28$ & $75.81\pm0.46$ & $75.90\pm0.57$ & \textbf{77.49}$\pm$ \textbf{0.36}\\
CMC & $73.46\pm0.35$ & $71.25\pm0.17$ & $51.94\pm0.78$ & $75.83\pm0.41$ & $74.36\pm0.21$ & \textbf{74.48}$\pm$\textbf{0.37} & $71.15\pm0.66$\\
Vowel & \textbf{96.91}$\pm$\textbf{0.64} & $96.11\pm0.19$ & $73.62\pm6.28$ & $95.36\pm0.21$ & $96.01\pm0.13$ & $96.43\pm0.24$ & $94.14\pm0.23$\\
Yeast4 & $94.29\pm0.25$ & $84.98\pm0.18$ & $87.06\pm1.73$ & $96.50\pm0.26$ & $75.10\pm0.16$ & $95.89\pm0.25$ & \textbf{96.36}$\pm$\textbf{0.14}\\
Shuttle-c0-vs-c4 & $97.36\pm1.69$ & $98.24\pm0.12$ & $82.14\pm2.19$ & $99.95\pm0.05$ & $99.84\pm0.12$ & $99.82\pm0.10$ & \textbf{99.98}$\pm$ \textbf{0.01}\\
Segment & $93.26\pm0.31$ & $92.79\pm0.35$ & $78.09\pm1.69$ & $96.48\pm0.17$ & $95.37\pm0.07$ & $95.76\pm0.42$ & \textbf{99.04}$\pm$ \textbf{0.24}\\
Abalone19 & $96.92\pm0.45$ & $81.63\pm0.94$ & $77.47\pm2.53$ & $84.12\pm1.55$ & $96.06\pm0.33$ & $87.02\pm0.89$ & \textbf{99.20}$\pm$\textbf{0.47}\\
Wisconsin & $96.72\pm0.73$ & $93.37\pm0.26$ & $93.10\pm0.24$ & $95.92\pm0.17$ & $95.34\pm0.25$ & $96.30\pm0.31$ & \textbf{97.21}$\pm$ \textbf{0.23}\\
Led7digit & $88.74\pm0.53$ & $87.23\pm0.62$ & $85.35\pm0.93$ & $90.78\pm0.08$ & \textbf{96.63}$\pm$\textbf{0.32} & $90.71\pm0.93$ & $91.85\pm0.43$\\

		\hline
\end{tabular*}
\end{center}
\end{table*}

\begin{table*}[!]
\caption {The testing results of linear classifiers (Mean G-mean($\%$)$\pm$ Standard Deviation)}
\label{Table.Gmlin}
\begin{center}
\begin{tabular*}{\textwidth}{l l l l l l l l }
	\hline
	Dataset & WSVM & $Over_{SVM}$ & $Under_{SVM}$ & $SMOTE_{SVM}$ & TWSVM & WLTSVM & FRLSTSVM\\
	\hline
Yeast3 & $78.02\pm1.20$ & $76.23\pm0.15$ & $71.75\pm1.69$ & $84.93\pm0.73$ & $83.07\pm0.96$ & $85.70\pm0.30$ & \textbf{87.32}$\pm$\textbf{0.61}\\
Vehicle & $90.38\pm0.74$ & $91.69\pm0.41$ & $86.91\pm0.80$ & \textbf{96.05}$\pm$\textbf{0.27} & $93.74\pm0.43$ & $95.48\pm0.56$ & $94.39\pm0.43$\\
Transfusion & $58.68\pm2.75$ & $61.31\pm2.72$ & $56.97\pm2.66$ & $ \textbf{62.06}\pm\textbf{3.18}$ & $52.72\pm0.36$ & $59.08\pm0.56$ & $59.98\pm0.82$\\
Wine & $76.57\pm1.64$ & $82.07\pm2.56$ & $80.20\pm2.73$ & $84.72\pm2.19$ & $86.87\pm1.36$ & $91.09\pm2.87$ & \textbf{95.90}$\pm$\textbf{0.15}\\
PimaIndian & $66.39\pm2.16$ & $71.24\pm0.92$ & $68.79\pm0.53$ & $72.01\pm1.51$ & $70.43\pm2.80$ & $73.16\pm1.42$ & \textbf{74.92}$\pm$\textbf{0.52}\\
Ionosphere & $75.59\pm1.26$ & $80.54\pm0.44$ & $62.09\pm0.41$ & $77.31\pm0.93$ & $75.67\pm0.97$ & $79.90\pm0.45$ & \textbf{89.60}$\pm$\textbf{0.15}\\
Haberman & $60.84\pm2.85$ & $62.90\pm0.67$ & $52.15\pm3.48$ & $66.92\pm0.50$ & $52.21\pm4.90$ & $64.72\pm2.77$ & \textbf{67.55}$\pm$\textbf{0.67}\\
CMC & $52.91\pm1.08$ & $50.58\pm2.17$ & $45.56\pm1.29$ & $60.43\pm1.76$ & $54.61\pm2.22$ & $63.44\pm0.34$ & \textbf{67.02}$\pm$\textbf{0.44}\\
Vowel & $75.48\pm2.42$ & $87.47\pm2.19$ & $64.06\pm3.50$ & $89.02\pm2.33$ & $85.62\pm1.51$ & \textbf{90.28}$\pm$\textbf{2.42} & $78.53\pm0.08$\\
Yeast4 & $73.84\pm2.01$ & $82.02\pm0.19$ & $67.93\pm1.97$ & $84.35\pm0.18$ & $18.78\pm6.20$ & \textbf{85.55}$\pm$\textbf{0.69} & $72.97\pm0.41$\\
Shuttle-c0-vs-c4 & $93.63\pm0.59$ & $99.51\pm0.12$ & $98.22\pm0.84$ & $98.95\pm0.17$ & $98.27\pm0.19$ & $99.01\pm0.39$ & \textbf{99.90}$\pm$\textbf{0.10}\\
Segment & $90.14\pm0.72$ & $89.70\pm1.74$ & $76.63\pm0.53$ & $94.79\pm1.48$ & $87.86\pm0.47$ & $93.64\pm1.11$ & \textbf{97.42}$\pm$\textbf{0.15}\\
Abalone19 & $62.75\pm3.82$ & $70.89\pm3.43$ & $57.39\pm2.36$ & $84.43\pm1.47$ & $8.097\pm.45$ & \textbf{85.35}$\pm$\textbf{1.91} & $79.23\pm0.34$\\
Wisconsin & $96.49\pm0.36$ & $91.40\pm0.21$ & $89.14\pm0.24$ & $93.54\pm0.18$ & $92.04\pm0.52$ & $95.93\pm0.35$ & \textbf{97.18}$\pm$\textbf{0.33}\\
Led7digit & $70.28\pm2.21$ & $66.41\pm1.67$ & $64.80\pm2.95$ & \textbf{73.48}$\pm$\textbf{2.04} & $62.56\pm5.34$ & $67.20\pm2.03$ & $69.23\pm0.33$\\
		\hline
\end{tabular*}
\end{center}
\end{table*}

\begin{table*}[!]
\caption {The testing results of nonlinear classifiers (Mean Accuracy($\%$)$\pm$ Standard Deviation)}
\label{Table.accNlin}
\begin{center}
\begin{tabular*}{\textwidth}{l l l l l l l l }

	\hline
	Dataset & WSVM & $Over_{SVM}$ & $Under_{SVM}$ & $SMOTE_{SVM}$ & TWSVM & WLTSVM & FRLSTSVM\\
	\hline
Yeast3 & $91.39\pm1.77$ & $92.91\pm1.04$ & $87.61\pm3.42$ & $93.52\pm1.49$ & $92.31\pm0.16$ & $93.54\pm0.15$ & \textbf{94.33}$\pm$\textbf{0.35}\\
Vehicle & $78.54\pm0.89$ & $80.63\pm1.89$ & $74.16\pm2.56$ & $82.80\pm1.52$ & $79.47\pm1.65$ & $82.09\pm2.18$ & \textbf{88.96}$\pm$\textbf{0.52}\\
Transfusion & $73.25\pm2.35$ & $74.85\pm1.38$ & $68.59\pm1.17$ & $75.60\pm2.59$ & $60.16\pm3.29$ & \textbf{76.99}$\pm$\textbf{0.47} & $76.62\pm0.87$\\
Wine & $84.28\pm2.18$ & \textbf{95.73}$\pm$\textbf{3.48} & $77.62\pm1.17$ & $92.51\pm2.64$ & $92.15\pm2.07$ & $94.67\pm2.12$ & $94.74\pm2.32$\\
PimaIndian & $62.79\pm1.61$ & $65.10\pm0.37$ & $64.42\pm2.28$ & $73.42\pm1.78$ & $72.23\pm1.27$ & $75.69\pm1.03$ & \textbf{76.9}$\pm$\textbf{0.61}\\
Ionosphere & $93.67\pm1.59$ & $91.63\pm2.64$ & $87.45\pm2.36$ & $94.32\pm1.71$ & $90.30\pm0.47$ & \textbf{95.27}$\pm$\textbf{2.29} & $79.68\pm1.12$\\
Haberman & $73.55\pm3.14$ & $74.24\pm1.15$ & $58.96\pm2.68$ & $72.77\pm2.41$ & $69.39\pm1.56$ & $73.48\pm0.35$ & \textbf{73.81}$\pm$\textbf{0.45}\\
CMC & $62.25\pm2.07$ & $68.13\pm1.65$ & $60.27\pm1.89$ & $71.74\pm2.20$ & \textbf{73.62}$\pm$\textbf{1.55} & $64.95\pm2.15$ & $70.13\pm1.52$\\
Vowel & $95.26\pm1.75$ & $91.80\pm3.14$ & $86.38\pm2.96$ & $96.29\pm2.33$ & $94.79\pm1.42$ & $97.58\pm1.35$ & \textbf{97.87}$\pm$\textbf{1.18}\\
Yeast4 & $92.63\pm1.39$ & $94.07\pm2.86$ & $92.49\pm1.73$ & $95.48\pm1.82$ & $92.16\pm2.79$ & $94.48\pm2.44$ & \textbf{96.22}$\pm$\textbf{1.06}\\
Shuttle-c0-vs-c4 & $92.79\pm1.88$ & $90.27\pm2.19$ & $86.47\pm3.51$ & $92.86\pm1.65$ & $92.07\pm0.35$ & $93.61\pm1.07$ & \textbf{99.81}$\pm$\textbf{0.04}\\
Segment & $49.53\pm2.83$ & $78.61\pm1.95$ & $46.09\pm1.69$ & $61.52\pm3.17$ & $82.59\pm1.87$ & $54.33\pm7.82$ & \textbf{99.42}$\pm$\textbf{0.33}\\
Abalone19 & $90.26\pm2.60$ & \textbf{93.74}$\pm$\textbf{1.82} & $85.73\pm1.45$ & $88.75\pm3.13$ & $92.29\pm2.27$ & $90.23\pm2.19$ & $81.42\pm1.18$\\
Wisconsin & $93.65\pm2.25$ & $94.26\pm0.86$ & $91.28\pm2.58$ & $96.16\pm0.81$ & $96.02\pm0.72$ & \textbf{96.96}$\pm$\textbf{0.22} & $93.12\pm0.51$\\
Led7digit & $89.28\pm1.67$ & $89.35\pm1.72$ & $86.76\pm1.85$ & $91.42\pm1.28$ & $90.99\pm2.81$ & $90.69\pm1.03$ & \textbf{96.18}$\pm$\textbf{0.41}\\

		\hline
\end{tabular*}
\end{center}
\end{table*}

\begin{table*}[!]
\caption {The testing results of nonlinear classifiers (Mean G-mean($\%$)$\pm$ Standard Deviation)}
\label{Table.Gmnlin}
\begin{center}
\begin{tabular*}{\textwidth}{l l l l l l l l }
	\hline
	Dataset & WSVM & $Over_{SVM}$ & $Under_{SVM}$ & $SMOTE_{SVM}$ & TWSVM & WLTSVM & FRLSTSVM\\
	\hline
Yeast3 & $81.05\pm1.46$ & $80.73\pm0.45$ & $74.14\pm1.25$ & $85.32\pm0.83$ & $78.68\pm2.57$ & $86.36\pm0.51$ & \textbf{88.63}$\pm$\textbf{1.04} \\
Vehicle & $81.05\pm1.39$ & $84.31\pm0.42$ & $69.41\pm2.44$ & $90.61\pm1.29$ & $83.08\pm3.30$ & \textbf{89.55}$\pm$\textbf{2.20} & $88.74\pm0.36$\\
Transfusion & $57.88\pm1.83$ & $50.89\pm2.89$ & $43.97\pm1.50$ & $60.52\pm2.04$ & $51.90\pm1.21$ & \textbf{66.46}$\pm$\textbf{1.15} & $64.44\pm0.57$\\
Wine & $7.52\pm18.94$ & $71.93\pm4.62$ & $0.00\pm0.00$ & $67.44\pm2.81$ & $60.87\pm2.68$ & $69.98\pm3.32$ & \textbf{91.51}$\pm$\textbf{1.45}\\
PimaIndian & $48.63\pm2.73$ & $54.89\pm3.11$ & $19.74\pm8.51$ & $58.42\pm3.20$ & $57.16\pm2.68$ & $60.50\pm4.02$ & \textbf{73.12}$\pm$\textbf{2.56}\\
Ionosphere & $93.93\pm1.60$ & $94.07\pm2.56$ & $72.44\pm2.96$ & $93.35\pm2.95$ & $90.73\pm3.14$ & \textbf{95.62}$\pm$\textbf{0.56} & $91.39\pm0.41$\\
Haberman & $35.52\pm1.47$ & $43.78\pm11.10$ & $29.26\pm2.18$ & $51.61\pm1.58$ & $41.37\pm2.36$ & $44.94\pm8.55$ & \textbf{63.21}$\pm$\textbf{2.58}\\
CMC & $64.04\pm7.38$ & $62.23\pm2.98$ & $54.34\pm2.00$ & $41.05\pm5.40$ & $53.64\pm2.19$ & $63.24\pm4.16$ & \textbf{69.54}$\pm$\textbf{4.28}\\
Vowel & $80.39\pm2.52$ & $90.24\pm3.55$ & $73.88\pm5.61$ & $93.39\pm4.34$ & $83.27\pm1.71$ & $95.66\pm1.61$ & \textbf{98.85}$\pm$\textbf{0.31}\\
Yeast4 & $31.13\pm3.01$ & $51.24\pm1.05$ & $40.17\pm4.15$ & $48.08\pm1.63$ & $46.66\pm2.54$ & $49.24\pm0.84$ & \textbf{61.59}$\pm$\textbf{1.29}\\
Shuttle-c0-vs-c4 & $62.07\pm2.09$ & $0.00\pm0.00$ & $58.14\pm2.96$ & $72.32\pm1.86$ & $69.87\pm3.19$ & $73.55\pm2.23$ & \textbf{99.62}$\pm$\textbf{0.13}\\
Segment & $58.27\pm3.10$ & $60.12\pm3.90$ & $38.75\pm4.28$ & $69.95\pm2.25$ & $67.03\pm2.66$ & $64.69\pm6.78$ & \textbf{99.35}$\pm$\textbf{0.04}\\
Abalone19 & $8.77\pm4.02$ & $12.17\pm2.75$ & $0.00\pm0.00$ & $17.51\pm3.11$ & $3.91\pm3.29$ & $9.21\pm0.35$ & \textbf{40.23}$\pm$\textbf{0.2}\\
Wisconsin & $75.98\pm0.53$ & $84.26\pm8.29$ & $90.48\pm2.51$ & $94.44\pm1.32$ & $93.85\pm3.20$ & \textbf{96.27}$\pm$\textbf{0.24} & $91.26\pm0.31$\\
Led7digit & $63.63\pm1.92$ & $56.01\pm1.44$ & $42.05\pm1.83$ & $68.03\pm2.62$ & $57.32\pm2.99$ & $66.10\pm2.17$ & \textbf{88.61}$\pm$\textbf{0.30}\\

		\hline
\end{tabular*}
\end{center}
\end{table*}

\begin{figure}[t]
	\centering
	\includegraphics[width=3.5in]{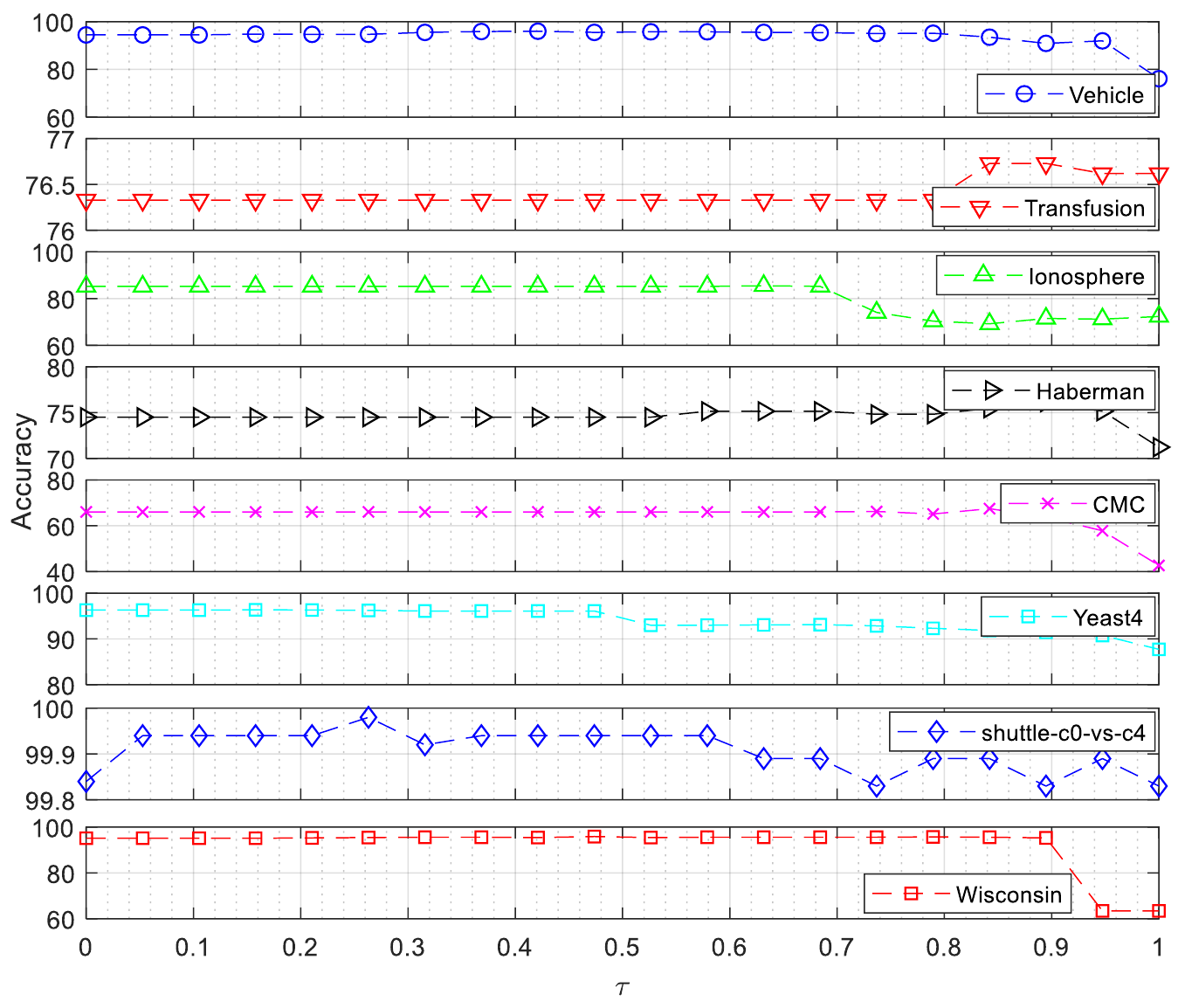}
	\caption{The behavior of the FRLSTSVM method by changing the parameter $\tau$.}
	\label{fig2}
\end{figure}

\begin{figure}[t]
	\centering
	\includegraphics[width=3.5in]{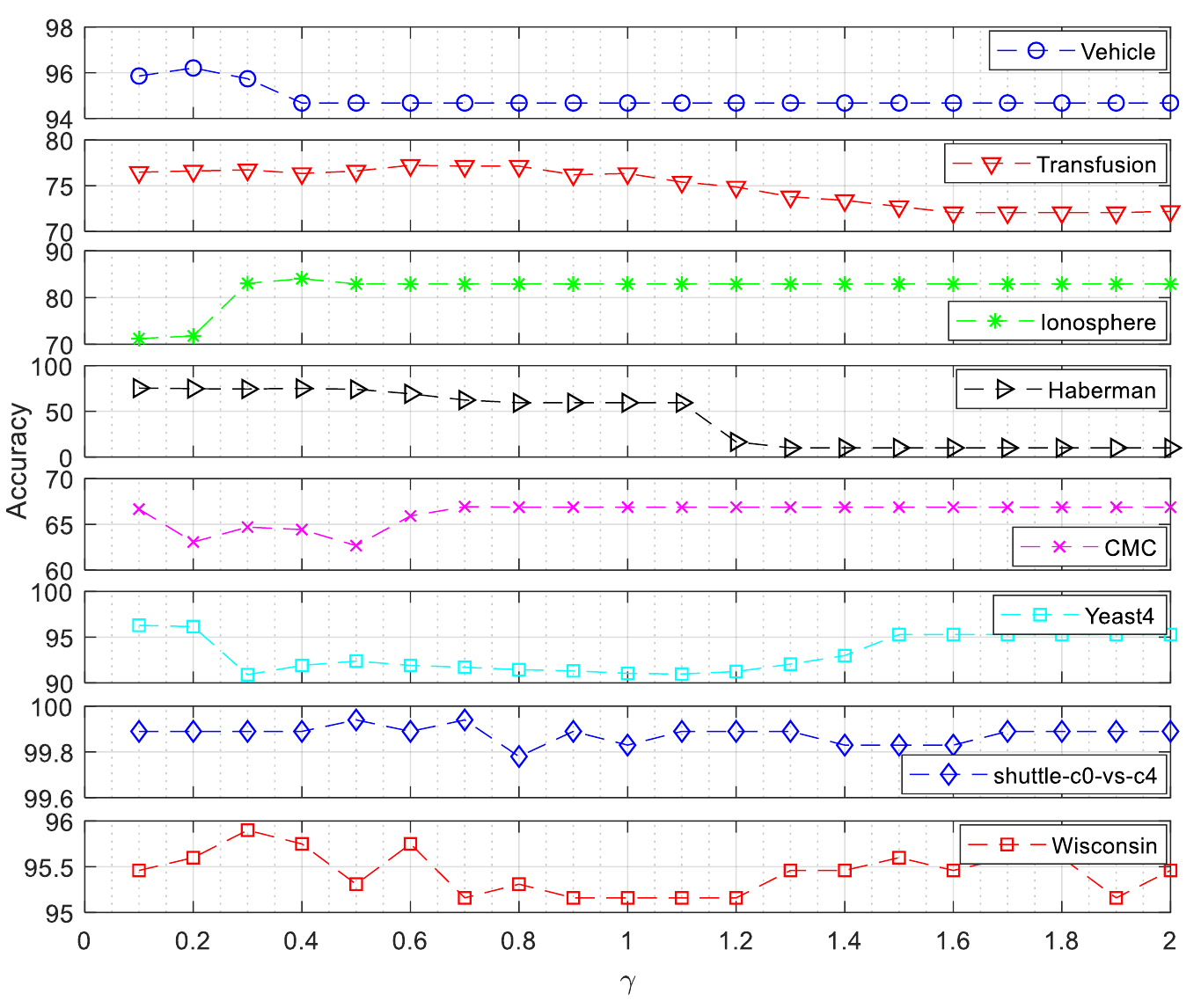}
	\caption{The behavior of the FRLSTSVM method by changing the parameter $\gamma$}
	\label{fig3}
\end{figure}

For the linear cases, Tables (\ref{Table.AccLin}) and (\ref{Table.Gmlin}) show that the proposed FRLSTSVM method achieves better performance in terms of G-mean metric compared to the previous methods in the most of benchmarks includes Yeast3, Wine, Pimaindian, Ionosphere, Haberman, CMC, Shuttle-c0-vs-c4, Segment, and Wisconsin. Also, it achieves better accuracy in 10 of all 15 datasets. Better performance of the proposed FRLSTSVM method in terms of G-mean metric on the most of datasets compare to other svm-based methods demonstrates better capability of FRLSTSVM method to trade-off between the sensitivity and specificity in the imbalanced problems.\par

The results in non-linear cases, show in Tables (\ref{Table.accNlin}) and (\ref{Table.Gmnlin}), are similar to those of the linear classifiers and demonstrate better performance of the FRLSTSVM method in terms of accuracy and G-mean metrics in the most of benchmarks includes Yeast3, Vehicle, Wine, Pimaindian, Haberman, CMC, Vowel, Yeast4, Shuttle-c0-vs-c4, Segment, Abalone19, and Led7digit datasets. In these datasets, compared to the WLTSVM, with the graph-based under-sampling strategy, the fuzzy-rough set strategy provides a better subsampling strategy by selecting an approprate set of the majority data points that represents the distribution of the data in the majority class and increases the robustness of classifier with respect to the outliers in those datasets.\par

Also, compared to TWSVM, the performance of FRLSTSVM in most of the datasets is better than TWSVM because there is not any bias in TWSVM for the majority or minority class in the imbalanced problem.

\subsection{Analysis of parameter}
\label{AnalysisOParameters}
The performance of the proposed FRLSTSVM method is highly dependent on two parameters, $\tau$ and $\gamma$, which must be tuned accurately. A non-negative threshold parameter $\tau$ is of membership type and must be tuned cross validation in the range $\{0, 0.05, 0.1, ... ,1\}$. The tuning of granularity parameter $\gamma$ used in fhe fuzzy relation (\ref{eq23}) is dependent on the dataset. We tune this parameter using cross validation in the range of $\{0.1, 0.2, ... ,2\}$. The behavior of the FRLSTSVM method on eight datasets by changing the parameters $\tau$ and $\gamma$ is show in figures (\ref{fig2}) and (\ref{fig3}), respectively.\par

We observe that the performance of the proposed method varies by changing these two parameters. For example, the proposed method for the Transfusion dataset obtains better performance by setting $\tau=0.8$ and $\gamma=0.6$. This means applying sub-sampling method among the the majority data points with at least 80\% similarity provides the best performance for this dataset. \par

Also, we apply the Gaussian kernel in the experiments of nonlinear FRLSTSVM method mentioned in the equations (\ref{eqNon1}) - (\ref{eqNon8}).

\section{Conclusions}
\label{7Conclusions}
In this paper, we had embedded weight biases in the TSVM formulations to overcome the bias phenomenon in the original TSVM and used fuzzy rough set theory to propose a new method based on the support vector machine for imbalanced data classification tasks. To have balanced training data, our method, called FRLSTSVM, introduces a new subsampling method based on the fuzzy-rough set theory that makes the classifier robust to the outliers and imbalanced data. In this strategy, the minority class data points remain unchanged, and a subset of the majority class data points are selected with a new method to reduce effectively the number of training points of the majority class. Also, to determine the weights in the formulation of FRLSTSVM, we have introduced a new strategy using fuzzy rough set theory.\par

The experimental results of the proposed FRLSTSVM method conducted on synthetic and real-world datasets demonstrate that this method is feasible and effective in terms of generalization ability and accuracy, for imbalanced data in most of the used benchmark datasets, compared to the number of related techniques.

%\begin{acknowledgements}
%\end{acknowledgements}


\bibliographystyle{spmpsci}      % mathematics and physical sciences

\bibliography{template}

\end{document}